\pdfoutput=1

\documentclass[twoside]{article}

\PassOptionsToPackage{sort&compress,numbers}{natbib}

\usepackage[accepted]{aistats2021}
\usepackage{natbib}

\usepackage[utf8]{inputenc} % allow utf-8 input
\usepackage[T1]{fontenc}    % use 8-bit T1 fonts
\usepackage{hyperref}       % hyperlinks
\usepackage{url}            % simple URL typesetting
\usepackage{booktabs}       % professional-quality tables
\usepackage{amsfonts}       % blackboard math symbols
\usepackage{nicefrac}       % compact symbols for 1/2, etc.
\usepackage{microtype}      % microtypography
\usepackage{numprint}       % for decimal places
\usepackage{lipsum}
\usepackage{amsmath}
\usepackage{dsfont}
\usepackage{float}
\usepackage{graphicx}
\graphicspath{{images/}{images/image_seg/}{images/controlled/}{images/imagenet21k/}}
\usepackage{caption}
\usepackage{subcaption}
\usepackage{subfiles}
\usepackage{placeins}
\usepackage{mathtools}
\usepackage{cuted}
\usepackage{listings,multicol}
\usepackage{tikz}
\usetikzlibrary{bayesnet}
\usetikzlibrary{arrows}

\usepackage{color}

\definecolor{mygreen}{rgb}{0,0.6,0}
\definecolor{mygray}{rgb}{0.5,0.5,0.5}
\definecolor{mymauve}{rgb}{0.58,0,0.82}
\allowdisplaybreaks

\DeclareMathOperator*{\argmax}{arg\,max}

\DeclareMathOperator*{\softmax}{softmax}
\newcommand{\brac}[1]{\left[#1\right]}

\newcommand{\parencurly}[1]{\left\{#1\right\}}

\newcommand\segw{0.95}

\newcommand\seg[1]{
\tiny{\noindent\hspace*{1.1cm} Image \hspace{1.45cm} Ground truth \hspace{1.3cm} Homoscedastic \hspace{0.63cm} Heteroscedastic $\tau = 1$ \hspace{0.25cm}
Heteroscedastic $\tau = 0.01$}
\includegraphics[width=\segw\textwidth]{appendix/seg_#1}
\noindent
\begin{minipage}{.15\textwidth}
\addtolength{\leftskip}{6 mm}
Heteroscedastic\\$\tau = 1$
\\[1.15cm]
Heteroscedastic\\ $\tau = 0.05$
\end{minipage}%
\begin{minipage}{.84\textwidth}
\includegraphics[height=0.3\textwidth]{appendix/uncert_#1}
\end{minipage}\newline\noindent\rule{\textwidth}{1pt}\vspace{3mm}}

\begin{document}



\twocolumn[

% \aistatstitle{Analysis of Softmax Approximation for Deep Classifiers under Input-Dependent Label Noise}
\aistatstitle{A Simple Probabilistic Method for Deep Classification under Input-Dependent Label Noise}

\aistatsauthor{ Mark Collier \And Basil Mustafa \And  Efi Kokiopoulou \And Rodolphe Jenatton \And Jesse Berent }

\aistatsaddress{ Google AI } ]

\begin{abstract}
  Datasets with noisy labels are a common occurrence in practical applications of classification methods. We propose a simple probabilistic method for training deep classifiers under input-dependent (heteroscedastic) label noise. We assume an underlying heteroscedastic generative process for noisy labels. To make gradient based training feasible we use a temperature parameterized softmax as a smooth approximation to the assumed generative process. We illustrate that the softmax temperature controls a bias-variance trade-off for the approximation. By tuning the softmax temperature, we improve accuracy, log-likelihood and calibration on both image classification benchmarks with controlled label noise as well as Imagenet-21k which has naturally occurring label noise. For image segmentation, our method increases the mean IoU on the PASCAL VOC and Cityscapes datasets by more than 1\% over the state-of-the-art model.
\end{abstract}

\section{Introduction}
\label{sec:background}

Public classification datasets are often designed to have clean labels \cite{deng2009imagenet}. However, when applying classification methods in practical settings one often has to deal with noisy labels. For example, large scale image classification datasets rely on images annotated by humans who may disagree over the correct label \cite{deng2009imagenet} or labels automatically generated from text surrounding an image on the web which may not match the image content \cite{li2017webvision}.
% \RJ{A two-liner example of a real-world setting with noisy labels would nicely warm-up the reader}.
This raises two practical problems; How to train deep classifiers in the presence of label noise? How to ensure that the trained model's predictions are well calibrated? We propose to answer both questions by constructing a principled probabilistic approach to modelling input-dependent label noise.

The uncertainty of a classification model can be divided into aleatoric and epistemic uncertainty \cite{kendall2017uncertainties}:
\begin{itemize}
    \item \textbf{Aleatoric uncertainty} captures inherent noise in the data. This uncertainty could be the result of noisy measurements, mis-labelled samples, unobserved predictive variables, and so on. Aleatoric uncertainty can be characterized as homoscedastic or heteroscedastic:
    \begin{itemize}
        \item \textbf{Homoscedastic}: the aleatoric uncertainty is constant across the input space.
        \item \textbf{Heteroscedastic}: the uncertainty varies across the input space, e.g.\ some samples may cause more disagreement amongst manual labellers than others.
    \end{itemize}
    \item \textbf{Epistemic uncertainty} captures uncertainty about the model that generated the data. This includes but is not limited to uncertainty over the parameters of the model.
\end{itemize}

The predictive uncertainty of a model is the combination of its aleatoric and epistemic uncertainty. In this paper we address the modelling of aleatoric uncertainty for classification tasks with noisy labels. Our approach can be combined with many approaches in the suite of Bayesian neural networks that estimate the epistemic uncertainty of a model \cite{gal2016dropout,gal2017concrete,blundell2015weight, neal1995bayesian, wilson2020bayesian, wenzel2020good} resulting in an estimate of the full predictive uncertainty.
% \RJ{(I have added a few refs)} \EK{Since we have the experiments ready for this, should we add them in the Appendix and mention this somewhere in the main text?}.
However, this is not the focus of this work. We note that epistemic uncertainty reduces to zero in the limit of infinite data, while aleatoric uncertainty is irreducible, so as datasets continue to increase in size, modeling aleatoric uncertainty will become increasingly important.

If a dataset contains heteroscedastic (i.e., input-dependent) label noise, then modelling heteroscedasticity is crucial for accurate uncertainty quantification and parameter estimation. Maximum likelihood estimation of a non-linear homoscedastic model on heteroscedastic data is biased and inconsistent \cite{greene2012econometric}. Thus, for datasets with such uncertainty, improved heteroscedastic modelling promises improved predictive performance with better calibrated predictions. The current best method for deep classifiers trained under heteroscedastic label noise, introduced by \citet{kendall2017uncertainties}, places a Normal distribution over the softmax logits and parameterizes the mean and variance of the Normal with deep neural networks.

% However it is unclear what the corresponding data generation process is to this method.\RJ{Right now, this is to me an unclear positioning and statement. More about this later.}

% \RJ{The related work seems extremely limited. What about all the other approaches tackling the problem of label noise (e.g., MentorNet etc.)? The paper does not have right now a clear related work section. It may be worth adding one at the end of the introduction. I would also connect to the literature that uses temperature scaling.
% Since the methodology proposed in the paper is arguably quite simple, we need to convince the readers that it has not been proposed yet. That's why a strong related work is important.  
% }

% The main contributions of this paper are:
In this paper, we make the following contributions:
\begin{enumerate}
    \item Inspired by the econometrics literature, we introduce a theoretical framework for deep heteroscedastic classification by viewing the use of the softmax as a smooth approximation to an argmax in the assumed data generation process. Interestingly, the state-of-the-art method \cite{kendall2017uncertainties} can be seen as a special case of our framework.
    \item Via this framework, we establish the importance of the softmax temperature in controlling a bias-variance trade-off for the approximation.
    \item We improve image classification and segmentation performance by tuning the softmax temperature. We compare to \citet{kendall2017uncertainties} and strong baselines from the noisy labels literature.
\end{enumerate}

\section{Background}
In order to motivate our development of heteroscedastic classification models we first review a heteroscedastic regression model by \citet{bishop1997regression} as it is particularly amenable to interpretation.

\subsection{Heteroscedastic Regression Models}
We have a dataset of examples: $\parencurly{(\mathbf{x}_1, y_1), ..., (\mathbf{x}_N, y_N)}$ where $y_i$ is real valued. We assume that $y_i$ are i.i.d. such that $y_i \sim \mathcal{N}(f^{\mathbf{w}}(\mathbf{x}_i), \sigma^{\mathbf{w}}(\mathbf{x}_i)^2)$, where $f^{\mathbf{w}}(\mathbf{x}_i)$ and  $\sigma^{\mathbf{w}}(\mathbf{x}_i)$ are parametric models parameterized by $\mathbf{w}$. The negative log-likelihood of the data is:
\begin{equation}
    \frac{1}{N} \sum_{i=1}^{N} \frac{1}{2 \sigma^{\mathbf{w}}(\mathbf{x}_i)^2} (y_i - f^{\mathbf{w}}(\mathbf{x}_i))^2 + \frac{1}{2} \log \sigma^{\mathbf{w}}(\mathbf{x}_i)^2.
\label{eq:hetero_regression_kendall}
\end{equation}
If we set $\sigma^{\mathbf{w}}(\mathbf{x}_i) = 1$ this reduces to a standard homoscedastic regression model. However for a non-constant function $\sigma^{\mathbf{w}}(\mathbf{x}_i)$, this differs from a standard regression model in that the squared error loss for each example is weighted by $1/{2 \sigma^{\mathbf{w}}(\mathbf{x}_i)^2}$. Those examples with higher predicted aleatoric uncertainty will be down-weighted in the learning objective, reducing overfitting to noisy labels.

% \RJ{Do we want to discuss the analogy with MentorNet?}

\subsection{Heteroscedastic Classification Models}

\citet{kendall2017uncertainties} extend this approach to the classification case. The Gaussian distribution is placed on the logits of a standard softmax classification model, making the logits latent variables:
\begin{equation}
\begin{split}
        &u_c \sim \mathcal{N}(f_c^{\mathbf{w}}(\mathbf{x}), \sigma_c^{\mathbf{w}}(\mathbf{x})^2), ~\forall c = 1...K, \\
        &p_c = \frac{\exp(u_c)}{\sum_{k=1}^K \exp(u_k)},
\end{split}
\label{eq:hetero_classification_kendall}
\end{equation}
where $p_c$ is the probability the label is class $c$ and $K$ is the number of classes. The model's log-likelihood is estimated by Monte Carlo (MC) sampling.

\section{Proposed Latent Variable Classification Model}
\label{sec:proposed_model}

\begin{figure*}
     \centering
     \begin{subfigure}[b]{0.45\textwidth}
         \centering
         \includegraphics[width=\textwidth]{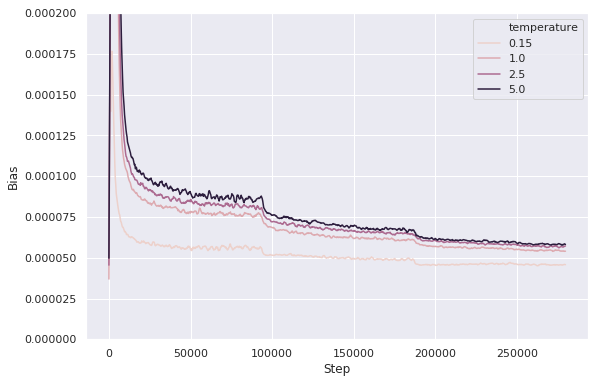}
     \end{subfigure}
     \begin{subfigure}[b]{0.45\textwidth}
         \centering
         \includegraphics[width=\textwidth]{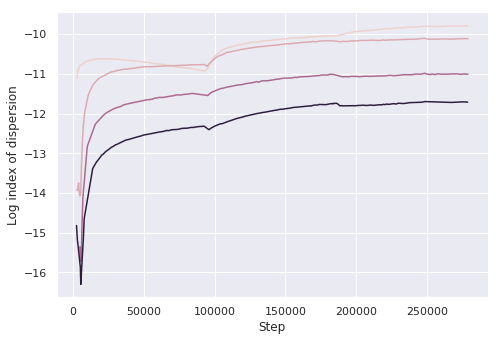}
     \end{subfigure}
        \caption{Bias-variance trade-off throughout training on Imagenet-21k.}
        \label{fig:bias_variance_imagenet21k}
\end{figure*}

\paragraph{Generative process}
We start by assuming a data generation process which is inspired by the econometrics literature and its associated terminology \cite{train2009discrete}.
%One way to motivate many standard classification models is to introduce a latent variable into the data generation process. We follow the econometrics literature in our terminology \cite{train2009discrete}, 
Suppose there is some latent utility $u_{c}$ associated with each class $c$. This utility is the sum of a deterministic reference utility $f_c^{\mathbf{w}}(\mathbf{x})$ and an unobserved stochastic component $\epsilon_c$. Class $c$ is chosen if its associated utility is greater than the utility for all other classes i.e.\ $y = c \Leftrightarrow u_{c} > u_{k}, \ \forall k \neq c$: 
% \RJ{I think it can be confusing to use $y$ both for the utility real-valued random variable and the class label discrete random variable. We can use, say, $u$ for the utility}
\begin{equation}
\begin{split}
    u_{c} &= f_c^{\mathbf{w}}(\mathbf{x}) + \epsilon_c \\
    p_c &= P(y = c | \mathbf{x}) = P(u_{c} > u_{k}, \ \forall k \neq c) \\
    &= P(\argmax_{k} u_{k} = c).
\end{split}
\label{eq:latent_variable_model}
\end{equation}
In the general case, the generative model in Eq.~(\ref{eq:latent_variable_model}) is intractable; however for certain choices of $\epsilon_c$ a closed-form solution is possible. For instance, assume $\epsilon_c$ is i.i.d.\ $\sim\mathit{G}(0,1)$, where $\mathit{G}$ is the Gumbel distribution, we then have:
\begin{equation}
\begin{split}
    p_c &= P(\argmax_{k} u_{k} = c) \\
    &= \mathbb{E}_{\epsilon_k \sim \mathit{G}(0,1)} \brac{\mathds{1} \parencurly{\argmax_{k}  u_{k} = c}} \\
    &= \frac{\exp(u_c)}{\sum_{k=1}^K \exp(u_k)}.
\end{split}
\label{eq:softmax_cross_entropy}
\end{equation}
$\mathds{1}$ is the 0-1 indicator function. The expectation in Eq.~(\ref{eq:softmax_cross_entropy}) has the closed form solution shown above \cite{train2009discrete}.

Interestingly, Eq.~(\ref{eq:softmax_cross_entropy}) is precisely the popular softmax cross-entropy model used in training neural network classification models. So standard neural network training is equivalent to assuming the latent variable generative process shown in Eq.~(\ref{eq:latent_variable_model}) with i.i.d.\ $\mathit{G}(0,1)$ noise.

%\subsection{Our Heteroscedastic Latent Variable Classification Model}

\paragraph{Breaking the i.i.d assumption}
The noise terms for standard softmax cross-entropy classification model, Eq.~(\ref{eq:softmax_cross_entropy}), are i.i.d., which thus implies a homoscedastic model. In order to be able to model heteroscedastic label noise we need to break this i.i.d.\ assumption. For our proposed heteroscedastic model we will assume the noise terms are independently but \textbf{not} identically distributed according to any location-scale distribution $\mathcal{D}$ (e.g., \ Gaussian or Gumbel): $\epsilon_c \sim \mathcal{D}(0, \sigma_c^{\mathbf{w}}(\mathbf{x})^2)$.

\paragraph{Estimating $p_c$ and its gradient}
Computing $p_c$ in Eq.~(\ref{eq:latent_variable_model}) requires computing an  expectation over $\epsilon_c$ which does not have a known analytic solution for a non-i.i.d.\ noise distribution \cite{train2009discrete}. We estimate this expectation via Monte Carlo (MC) sampling.
The MC estimate's derivatives are either zero or undefined due to the $\argmax$ in the generative process. Therefore we seek a smooth approximation to the $\argmax$ in Eq.~(\ref{eq:latent_variable_model}). Similar to the development of the Gumbel-Softmax \cite{gumbelsoftmax2017,concrete2017}, we note that in a zero temperature ($\tau$) limit the softmax function is equivalent to the argmax, hence:
\begin{equation}
\begin{split}
    p_c &= \mathbb{E}_{\mathbf{\epsilon} \sim \mathcal{D}(0, \mathbf{\sigma^{\mathbf{w}}}(\mathbf{x})^2)} \brac{\mathds{1} \parencurly{\argmax_{k} u_{k} = c}} \\
    &= \mathbb{E} \brac{\lim_{\tau \to 0} \frac{\exp(u_{c}/\tau)}{\sum_{k=1}^K \exp(u_{k}/\tau)}} \\
    &\approx \mathbb{E} \brac{\frac{\exp(u_{c}/\tau)}{\sum_{k=1}^K \exp(u_{k}/\tau)}}, \ \tau > 0,
\end{split}
\label{eq:p_c_softmax}
\end{equation}
where the expectation is still over $\epsilon_k \sim \mathcal{D}(0, \sigma_k^{\mathbf{w}}(\mathbf{x})^2)$.
The softmax is a smooth approximation to the assumed generative model. The approximation is exact in a zero temperature limit but biased and differentiable for positive $\tau$. A similar result for binary classification with sigmoid smoothing function is derived in Appendix \ref{sec:binary_classification}.

 In order to compute gradients w.r.t.\ our stochastic model, we apply the reparametrization trick \cite{diederik2014auto, rezende2014stochastic} and rewrite $u_c$ as a \textit{deterministic} function of $\mathcal{D}(0, 1)$. A reparameterized MC estimate of the approximate predictive probabilities $p_c$ can be obtained as:
\begin{equation}
\begin{split}
    p_c &\approx \frac{1}{S} \sum_{s=1}^{S} \frac{\exp((f_c^{\mathbf{w}}(\mathbf{x}) + \sigma_c^{\mathbf{w}}(\mathbf{x}) \epsilon_c^s)/\tau)}{\sum_{k=1}^K \exp((f_k^{\mathbf{w}}(\mathbf{x}) + \sigma_k^{\mathbf{w}}(\mathbf{x}) \epsilon_k^s)/\tau)}, \\ 
    &\epsilon_1^s, ..., \epsilon_K^s \sim \mathcal{D}(0, 1),
\end{split}
\label{eq:p_c_softmax_mc}
\end{equation}
where $S$ is the number of MC samples. In actual implementation, we typically compute $f_c^{\mathbf{w}}(\mathbf{x})$ and $\sigma_c^{\mathbf{w}}(\mathbf{x})$ as a linear function of a shared representation of $\mathbf{x}$ outputted by the final layer of a neural network. Computing Eq.~(\ref{eq:p_c_softmax_mc}), with $S$ samples, has computational complexity $\mathcal{O}(SK + DK)$. This is typically trivial relative to the complexity of computing the shared representation.

\paragraph{Bias-variance trade-off}
In our softmax approximation, as the temperature $\tau$ gets closer to zero, the bias in the approximation to the true objective goes down, but the variance of the MC estimate of the gradients of the approximate objective increases \cite{gumbelsoftmax2017,concrete2017}. Thus, the temperature parameter controls a bias-variance trade-off. We test this claim empirically on the Imagenet-21k dataset, see Fig.\ \ref{fig:bias_variance_imagenet21k} which shows this trade-off throughout training. As expected log index of dispersion, a normalized metric for gradient variance, is higher at lower temperatures while bias is lower at lower temperatures. This effect persists throughout training.

We follow the existing literature in computing the gradient variance using an exponential moving average of the first and second moments of the gradient vector \cite{tucker2017rebar}. However we report the log index of dispersion which normalizes the gradient variance by the absolute value of the expected gradient. This accounts for the effect that the magnitude of the gradient vector varies a lot throughout training. The log index of dispersion is computed as $\log \hat{\sigma}^2/| \hat{\mu}|$ where $\hat{\sigma}^2$ is the estimated gradient variance and $\hat{\mu}$ is the estimated expected gradient. We compute the metric element-wise for each trainable parameter and average over all parameters.

Bias is measured as the expected KL divergence between the hard argmax samples and softmax samples at a given temperature i.e.\ $ \textrm{bias} = \mathbb{E}_{\epsilon \sim \mathcal{D}(0, \sigma(\mathbf{x})^2)} \brac{D_{KL}(\argmax \mathbf{u} || \softmax_{\tau} \mathbf{u})}$.

\paragraph{Connecting with existing methods}
Interestingly, the method of \citet{kendall2017uncertainties}, Eq.~(\ref{eq:hetero_classification_kendall}) can be seen a special case of our framework by setting $\tau = 1.0$ and choosing a Gaussian noise distribution. However, differently from our method, the softmax in the model of \cite{kendall2017uncertainties} does not show up as an argmax approximation in the generative process and therefore the role of the temperature parameter in the training dynamics is not recognized. We have shown above that the temperature does in fact control a bias-variance trade-off that persists throughout training. In the experiments below, we shall show that the effect of the temperature parameter on the training dynamics has a significant impact on the test set model performance.

Note in passing that Platt-scaling/temperature scaling \cite{platt1999probabilistic,guo2017calibration} is a simple and popular post-hoc calibration method for neural network classifiers, which may look similar to our method at a first glance. This method tunes the softmax temperature parameter on a validation set \emph{after training} the neural network in order to improve test set calibration and log-likelihood. However, unlike our approach which alters the model training dynamics, this method \emph{cannot} affect the model accuracy (because the temperature does not change the maximum of the softmax function). In the experimental section we demonstrate empirically that our method improves accuracy and can be also successfully combined with post-hoc temperature scaling.

Note finally that a similar approximation has been studied in the econometrics literature, where it is known as the logit smoothed accept-reject simulator \cite{train2009discrete,mcfadden1989method,bolduc1996multinomial}. However the econometrics treatment is restricted to linear models. The latent variable formulation of heteroscedastic classification models is also standard in the Gaussian Processes literature where it is assumed the latent noise $\epsilon_c$ is distributed Gaussian \cite{hernandez2014mind,williams2006gaussian} and a GP prior is placed on $f(\mathbf{x})$ and $\log \sigma(\mathbf{x})^2$. Again exact inference on the likelihood is intractable and different approximate inference methods are used \cite{hernandez2014mind}.

\section{Related Work}
\label{sec:related_work}

\subsection{Aleatoric Uncertainty in Deep Learning}

Estimating uncertainty in deep learning has mostly focused on epistemic uncertainty \cite{gal2016dropout,gal2017concrete,blundell2015weight, neal1995bayesian, wilson2020bayesian, wenzel2020good}. Nevertheless, for heteroscedastic regression \citet{bishop1997regression} were early proponents of parameterizing the mean and variance term in a Gaussian likelihood with neural networks.

Follow-up work \cite{kendall2017uncertainties} revisits this regression model and introduces the heteroscedastic classification model discussed earlier in this paper. The authors show that these heteroscedastic models can be combined with MC dropout \cite{gal2016dropout} approximate Bayesian inference for epistemic uncertainty estimation. The combined heteroscedastic Bayesian model yields improved performance on semantic segmentation and depth regression tasks. \citet{deepensembles.2018} propose an ensembling approach to uncertainty estimation in deep learning using multiple models to estimate both aleatoric and epistemic uncertainty. Along the same lines \citet{liu2019accurate} introduce a Bayesian non-parametric ensemble to estimate both sources of uncertainty. \citet{ayhan2018testtimeDA} propose estimating heteroscedastic aleatoric uncertainty by measuring the variation in the network's output under standard data augmentation.

Some recent research efforts aim to estimate specific uncertainty metrics. \citet{pearce2018predictionintervals} introduce a novel loss function, which allows them to use ensemble networks to estimate prediction intervals without making any assumptions on the output distribution. \citet{natasa2019uncertainties} introduce a quantile regression loss function in order to simultaneously learn all the conditional quantiles that are subsequently used to compute well-calibrated prediction intervals.

\begin{figure*}[tbh]
    \centering
    \includegraphics[width=0.95\textwidth]{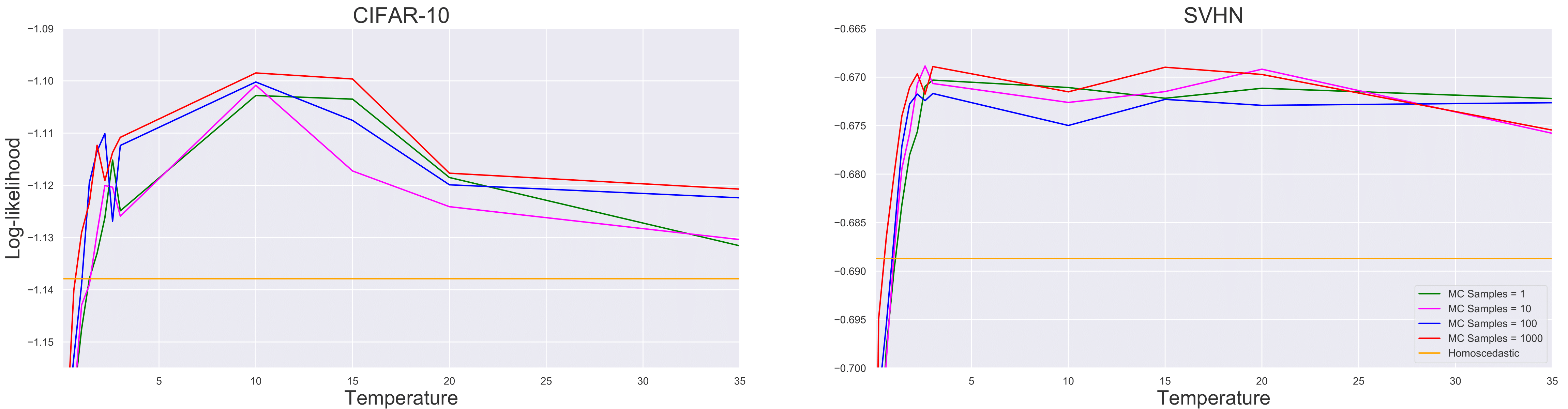}
    \caption{Effect of temperature and number of training MC samples on \textit{noisy} test set log-likelihood.}
    \label{fig:synthetic_results}
\end{figure*}

\subsection{Noisy Labels}
\label{sec:related_work_noisy_labels}

A large literature exists, which seeks to tackle the problem of classification with noisy labels using deep neural networks. Most of the methods try to identify samples with incorrect labels and remove or under-weight these samples in the loss function. 
Bootstrapping \cite{reed6596training} attempts to denoise the labels by setting the target label to be a linear combination of the (potentially noisy) label and the current model's predictions.
The MentorNet method \cite{MentorNet.2018} introduces a second neural network, the MentorNet, which estimates a curriculum learning strategy of weighting examples for training a StudentNet (i.e., the main network). The MentorNet can be learned to approximate a pre-defined curriculum or discover a new curriculum from data. In the latter case the curriculum is learned using a small dataset with clean labels. MentorMix \cite{jiang2020beyond} combines MentorNet with \textit{mixup} regularization \cite{zhang2018mixup}. The Co-teaching method \cite{CoTeaching.2018} also jointly trains two neural networks. At each training step, both networks compute predictions on a mini-batch of samples and identify small loss samples, which are then fed to the other network for learning. The underlying assumption is that small loss examples are more likely to have clean labels. In Appendix \ref{sec:mentornet_connection} we show that in the regression case the MentorNet objective is equivalent to the heteroscedastic regression objective, Eq.~(\ref{eq:hetero_regression_kendall}).

Note that our method can be applied to this problem of classification with noisy labels and we provide empirical comparisons with these methods in the next section. Differently from these methods, it is important to emphasize that our method also provides estimates of aleatoric uncertainty, which for some applications (e.g., autonomous driving), may be an object of interest in its own right. 
%Unlike some methods in the noisy labels literature, 
Our method can be also \emph{naturally} combined with Bayesian methods for epistemic uncertainty estimation.

\section{Experiments}
\label{sec:experiments}

In real-world applications of machine learning noisy labelled datasets are the norm \cite{li2017webvision}, however public classification datasets are typically collected in such a manner as to avoid noisy labels \cite{deng2009imagenet}. In the below experiments we first evaluate our method on two image classification datasets (CIFAR-10 \cite{krizhevsky2009learning} and SVHN \cite{netzer2011reading}) where we generate heteroscedasticity synthetically. Next, we apply our method to Imagenet-21k \cite{deng2009imagenet} a large scale image classification dataset with naturally occurring label noise for which we do not have to introduce synthetic label noise. We further evaluate our method on two image segmentation benchmarks which also exhibit heteroscedasticity naturally \cite{kendall2017uncertainties}, PASCAL VOC \cite{everingham2014pascal} and Cityscapes \cite{cordts2016cityscapes}. For all experiments, we choose the Gaussian noise distribution in our method for comparability to \citet{kendall2017uncertainties}.

\subsection{Controlled Label Noise}
\label{sec:synthetic_experiments}

\npdecimalsign{.}
\npfourdigitnosep
\nprounddigits{3}
\begin{table}[tbh]
\setlength{\tabcolsep}{1.3pt}
\caption{CIFAR-10 and SVHN test dataset performance. Negative Log-likelihood (NLL) and expected calibration error (ECE) \cite{guo2017calibration} are computed on the noisy test set (with the same label corruption process applied to the training set). Clean accuracy (Acc) is computed on the test set with uncorrupted labels. For our method, optimal $\tau^\ast$ is determined by validation set log-likelihood. Number of samples $S = 1000$. $p$-values are from a paired sample two-tailed t-test with 25 replicates from corresponding random seeds. T-tests are conducted in reference to our method.}
\label{table:hetero_vs_homo}
\centering
\begin{tabular}{lcccccc}
\toprule
Method &
  \multicolumn{3}{c}{CIFAR-10 ($\tau^\ast = 3$)} &
  \multicolumn{3}{c}{SVHN ($\tau^\ast = 10$)} \\
 & NLL & Acc & ECE & NLL & Acc & ECE \\
  \midrule
Homoscedastic &
  $\numprint{1.1378800120000003}^{\ddagger}$ &
  $\numprint{0.8427520000000003}^{\ddagger}$ &
  $\numprint{0.04342196612}$ &
  $\numprint{0.6886919395999999}^{\ddagger}$ &
  $\numprint{0.950030734}^{\ddagger}$ &
  $\numprint{0.01811325112}^{\dagger}$ \\
% \ \ + Platt-scaling &
%   $\numprint{1.1242398320000002}^{\ddagger}$ &
%   $\numprint{0.8427520000000003}^{\ddagger}$ &
%   $\numprint{0.030100399799999997}^{\ddagger}$ &
%   $\numprint{0.6856170852}^{\ddagger}$ &
%   $\numprint{0.950030734}^{\ddagger}$ &
%   $\numprint{0.01523184818}^{\ddagger}$ \\
$\tau = 1.0$ \cite{kendall2017uncertainties} &
  $\numprint{1.1290426839999999}^{\ddagger}$ &
  $\numprint{0.853088}^{\dagger}$ &
  $\numprint{0.049608294080000004}^{*}$ &
  $\numprint{0.6800222575999999}^{\ddagger}$ &
  $\numprint{0.9517240360000001}^{\ddagger}$ &
  $\numprint{0.014971369119999998}$ \\
\ + Platt.\ &
  $\numprint{1.111916096}^{\ddagger}$ &
  $\numprint{0.853088}^{\dagger}$ &
  $\numprint{0.02576470008}^{*}$ &
  $\numprint{0.6780429372000001}^{\ddagger}$ &
  $\numprint{0.9517240360000001}^{\ddagger}$ &
  $\numprint{0.013134200620000001}^{*}$ \\
Ours $\tau = \tau^\ast$ &
  $\mathbf{\numprint{1.098491052}}$ &
  $\mathbf{\numprint{0.858624}}$ &
  $\mathbf{\numprint{0.04191602044000001}}$ &
  $\mathbf{\numprint{0.6689119888}}$ &
  $\mathbf{\numprint{0.9554671156000001}}$ &
  $\mathbf{\numprint{0.013474776364000003}}$ \\
\ + Platt.\ &
  $\mathbf{\numprint{1.0862124160000004}}$ &
  $\mathbf{\numprint{0.858624}}$ &
  $\mathbf{\numprint{0.021780285459999996}}$ &
  $\mathbf{\numprint{0.6688605748}}$ &
  $\mathbf{\numprint{0.9554671156000001}}$ &
  $\mathbf{\numprint{0.0110610903}}$ \\
  \midrule
Bootstrapping &
  $\numprint{1.1569313920000002}^{\ddagger}$ &
  $\numprint{0.8509359999999999}^{\ddagger}$ &
  $\numprint{0.07696113196}^{\ddagger}$ &
  $\numprint{0.6879735243999999}^{\ddagger}$ &
  $\numprint{0.9527719783999999}^{\ddagger}$ &
  $\numprint{0.03523115396}^{\ddagger}$\\    
% \ \ + Platt-scaling &
%   $\numprint{1.1215284280000002}^{\ddagger}$ &
%   $\numprint{0.8509359999999999}^{\ddagger}$ &
%   $\numprint{0.034695889439999995}^{\ddagger}$ &
%   $\numprint{0.6776840907999998}^{\ddagger}$ &
%   $\numprint{0.9527719783999999}^{\ddagger}$ &
%   $\mathbf{\numprint{0.011247024875999998}}$ \\
MentorNet &
  $\numprint{1.7775430679999997}^{\ddagger}$ &
  $\numprint{0.8576960000000001}$ &
  $\numprint{0.14402864735999998}^{\ddagger}$ &
  $\numprint{1.3733280599999997}^{\ddagger}$ &
  $\mathbf{\numprint{0.9554931212000001}}$ &
  $\numprint{0.10695505087999999}^{\ddagger}$ \\
% \ \ + Platt-scaling &
%   $\numprint{1.5958343599999998}^{\ddagger}$ &
%   $\numprint{0.8576960000000001}$ &
%   $\numprint{0.1326711428}^{\ddagger}$ &
%   $\numprint{1.0503218928}^{\ddagger}$ &
%   $\mathbf{\numprint{0.9554931212000001}}$ &
%   $\numprint{0.09570506232000003}^{\ddagger}$ \\
Co-teaching &
  $\numprint{2.506220476}^{\ddagger}$ &
  $\numprint{0.8568159999999999}$ &
  $\numprint{0.21968032959999997}^{\ddagger}$ &
  $\numprint{1.9085762600000005}^{\ddagger}$ &
  $\numprint{0.9541425995999999}$ &
  $\numprint{0.151293794}^{\ddagger}$ \\
% \ \ + Platt-scaling &
%   $\numprint{2.157522484}^{\ddagger}$ &
%   $\numprint{0.8568159999999999}$ &
%   $\numprint{0.20121231439999998}^{\ddagger}$ &
%   $\numprint{1.3556756719999996}^{\ddagger}$ &
%   $\numprint{0.9541425995999999}$ &
%   $\numprint{0.13891620064000001}^{\ddagger}$ \\
\bottomrule
\end{tabular}
\begin{flushleft}
\small
 $^*$ p < 0.05 \hspace{5mm}
 $^\dagger$ p < 0.01  \hspace{5mm}
 $^\ddagger$ p < 0.001
\end{flushleft}
\end{table}

We generate heteroscedasticity synthetically in two standard image classification datasets; CIFAR-10 \cite{krizhevsky2009learning} and SVHN \cite{netzer2011reading}. We corrupt the labels of examples in a data-conditional manner as follows: labels 1-4 are not corrupted, label 5 is flipped 10\% of the time, label 6 20\%, proceeding in 10\% increments to 60\% for label 10. We use the same architecture as a baseline from the noisy labels literature \cite{CoTeaching.2018}, see Appendix \ref{sec:synthetic_appendix} for details.

Fig.\ \ref{fig:synthetic_results} shows the test set log-likelihood as a function of the softmax temperature, averaged over 25 training runs. The number of MC samples during training is varied. However when making predictions on the test set and the validation set we always use 1,000 samples. The plots show a characteristic curve of a bias-variance trade-off, confirming the role and importance of the softmax temperature. Our method is also robust to the number of MC samples during training. See Appendix \ref{sec:more_synthetic_results} for similar plots for test set accuracy. Table \ref{table:hetero_vs_homo} shows the log-likelihood and expected calibration error on the noisy test set and accuracy on the clean test set for all methods including baselines from the noisy labels literature. In what follows, we further discuss the results shown in Fig.\ \ref{fig:synthetic_results} and Table \ref{table:hetero_vs_homo}.

\paragraph{\textbf{Do heteroscedastic models outperform homoscedastic models when there exists heteroscedastic noise?}}

First we wish to verify whether in fact heteroscedastic models outperform the standard homoscedastic model when we know there exists heteroscedastic noisy labels. Looking at Fig.\ \ref{fig:synthetic_results} it is clear that there are large ranges of temperatures for which the heteroscedastic test set log-likelihood is higher than the homoscedastic model. This is true for all numbers of training set MC samples. See Fig.\ \ref{fig:synthetic_results_acc} in Appendix \ref{sec:more_synthetic_results} for similar plots for test set accuracy.

We also conduct a more formal test. We select the optimal temperature for each dataset based on the \textit{validation set} log-likelihood. Then we conduct a paired sample t-test between the homoscedastic model and our heteroscedastic model at the optimal temperature on the \textit{test set}, with $S = 1000$. Replicates in the t-test are paired by having corresponding random seeds. In Table \ref{table:hetero_vs_homo} we see that for each dataset the best heteroscedastic model does in fact outperform the homoscedastic model and that the difference in test set log-likelihood and accuracy is statistically significant.

\paragraph{\textbf{Is 1.0 always the optimal $\tau$?}}

We compare our method to \citet{kendall2017uncertainties} who implicitly set the softmax temperature to $1.0$. We perform a paired t-test between the heteroscedastic model at optimal temperature (our method) and at $\tau = 1.0$ on the test set. Table \ref{table:hetero_vs_homo} shows that the optimal temperature is greater than $1.0$ on both datasets and the difference in log-likelihood between the optimal temperature and $\tau = 1.0$ is statistically significant on all datasets. Thus the optimal temperature is not always $1.0$ and the performance of heteroscedastic models can be improved by tuning the softmax temperature. Image segmentation and Imagenet-21k results reported below confirm this.

\paragraph{\textbf{Noisy labels baselines}}

We implement three baselines from the noisy labels literature; Co-teaching \cite{CoTeaching.2018}, Self-Paced MentorNet \cite{MentorNet.2018} and Bootstrapping \cite{reed6596training}. These methods are reviewed in \S \ref{sec:related_work_noisy_labels} and implementation details are provided in Appendix \ref{sec:synthetic_appendix}. Co-teaching and MentorNet do not provide calibrated predictions on the test set (even with Platt-scaling), additionally our method provides better or equal accuracy than each noisy labels baseline on both datasets.

Under our input-dependent label noise generation process, our method outperforms the previous state-of-the-art method for heteroscedastic aleatoric uncertainty modelling for deep classifiers \textit{and} the state-of-the-art methods for training deep classifiers with noisy labels.

\paragraph{\textbf{Can post-hoc calibration reverse the gains from our method?}} 

Platt-scaling/temperature scaling \cite{platt1999probabilistic,guo2017calibration} is a method for post-hoc calibration of classifiers. After training we optimize the softmax temperature on the validation set log-likelihood. Due to space constraints we present the effect of Platt-scaling only on our method and the method of \citet{kendall2017uncertainties} in Table \ref{table:hetero_vs_homo}. This ablation is presented for all methods in Appendix \ref{sec:platt_scaling}. Platt-scaling improves all methods test-set log-likelihood and ECE, including our method. The combination of our method with post-hoc calibration yields the best results. Unlike our method, Platt-scaling has no effect on the training dynamics and no effect on the accuracy of the model.

\paragraph{Additional results}
Due to lack of space, we provide additional results with controlled label noise in appendices \ref{sec:no_noise}, \ref{sec:uniform_noise} and \ref{sec:varying_noise}. These are our main findings:
\begin{itemize}
    \item Our model results in improved test set log-likelihood, accuracy and calibration on the original CIFAR-10 and SVHN datasets, \textit{even when no noise is added to the labels} (Appendix \ref{sec:no_noise}). 
    \item We show the utility of our model when the controlled label noise is homoscedastic (Appendix \ref{sec:uniform_noise}) 
    \item We validate that as we increase the level of heteroscedastic noise our model provides increasing improvements over the baselines (Appendix \ref{sec:varying_noise}).
\end{itemize}

\begin{figure}
    \centering
    \includegraphics[width=0.475\textwidth]{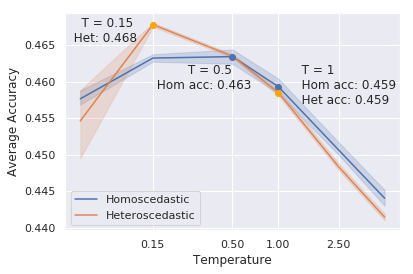}
    \caption{Imagenet-21k: Average accuracy vs.\ temperature, averaged over 5 random seeds. The shaded area shows bootstrapped 95\% confidence intervals.}
    \label{fig:acc_vs_temp_imagenet21k}
\end{figure}

\begin{figure*}[tbh]
     \centering
     \begin{subfigure}[b]{\textwidth}
     \centering
     \includegraphics[width=\textwidth]{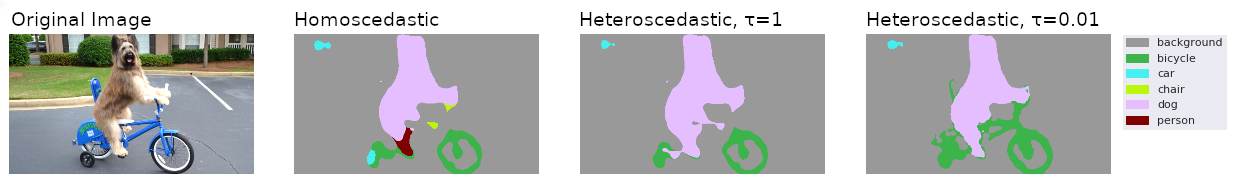}
     \caption{Image segmentation for an image from the internet (not in the PASCAL VOC dataset).
     }
     \label{fig:dog_segs}
     \end{subfigure}
     \begin{subfigure}[b]{\textwidth}
     \centering
     \includegraphics[width=\textwidth]{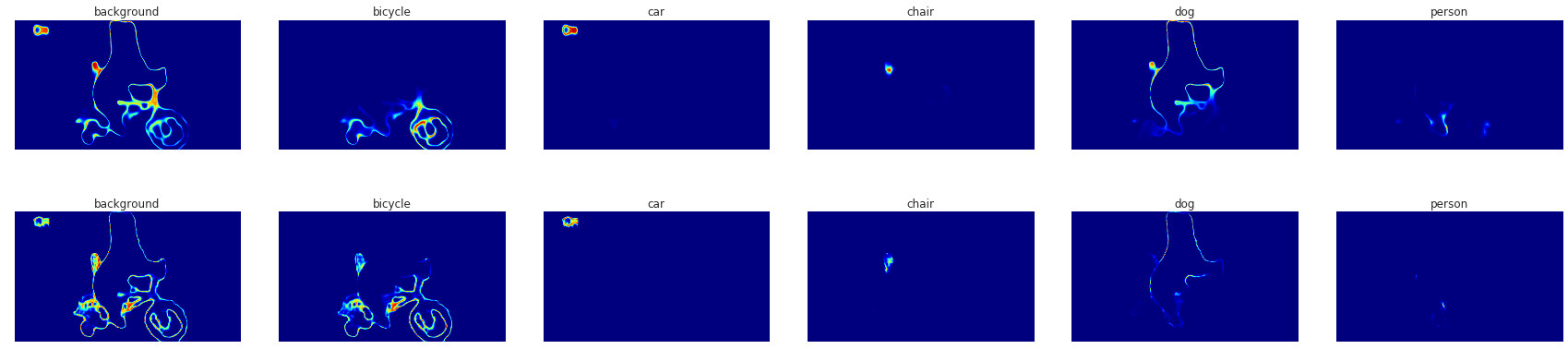}
     \caption{Aleatoric uncertainty heatmaps for heteroscedastic models: $\tau=1.0$ (top) and optimal temperature $\tau^*=0.05$ (bottom).
     }
     \label{fig:uncertainty}
     \end{subfigure}
     \caption{Example image segmentation and per class predicted variance heatmap.}
\end{figure*}

\subsection{Imagenet-21k}
\label{sec:imagenet21k}

Imagenet-21k is a larger version of the standard ILSVRC-2012 Imagenet benchmark \cite{deng2009imagenet,kolesnikov2019big,beyer2020we}. It has over 12.8 million images with 21,843 classes. Images may have multiple classes. We train a Resnet-152 \cite{he2016deep} for 90 epochs. The initial learning rate is 0.1 and is reduced by a factor of 10 after epoch 30, 60 and 80. We apply l2 regularization with penalty $3 \times 10^{-3}$. We use stochastic gradient descent with momentum factor 0.9 for optimization. No standard train/test split is provided, so we use 4\% of the dataset as a validation set and a further 4\% as the test set.

\begin{table}[tbh]
\centering
\caption{Imagenet-21k results for heteroscedastic and homoscedastic models. The test set average accuracy (Avg Acc) over the 21,843 classes $\pm$ 1 standard deviation is reported. $^\ddagger$ indicates that an unpaired t-test between the method and our method had p-value < 0.001. 5 runs from different random initializations are used. }
\label{table:het_vs_hom_imagenet21k}
\begin{tabular}{lc}
\toprule
Method & Avg Acc \\
\midrule
Homoscedastic $\tau = 1.0$ & 0.459$^\ddagger$ ($\pm 0.0014$) \\
Homoscedastic $\tau^\ast = 0.5$ & 0.463$^\ddagger$ ($\pm 0.0012$) \\ 
Heteroscedastic $\tau = 1.0$ \cite{kendall2017uncertainties} & 0.459$^\ddagger$ ($\pm 0.0006$) \\ 
Ours $\tau^\ast = 0.15$ & \textbf{0.468}  ($\pm 0.0004$)\\ 
\bottomrule
\end{tabular}
\end{table}

%\EK{Up to this point we haven't mentioned anything about varying the temperature in the homoscedastic model and why we would want to do that. It will be good to explain why we do that in the previous section where we give an overview of the experiments.}

Table \ref{table:het_vs_hom_imagenet21k} shows that our heteroscedastic model provides more accurate predictions than the baselines. To test the significance of the results we conducted an unpaired two-tailed t-test between our method and each baseline over 5 random seeds. For completeness, we also report the standard deviation of the average accuracy. Our method improves average accuracy by 0.9\% over standard neural network training and the heteroscedastic model at $\tau = 1.0$. Note that the model of \citet{kendall2017uncertainties} does not provide any improvement over the homoscedastic model in this dataset.

We test whether the improvement from our method is simply due to tuning another hyperparameter or if it is caused by the temperature playing a unique role for the heteroscedastic model. To conduct this test, we also tune the softmax temperature \emph{during training} for the homoscedastic model. For the homoscedastic model, the softmax temperature controls the scaling of the random initialization of the final layer weights. As we would expect, tuning this hyperparameter improves performance, but the effect size is much smaller than for the heteroscedastic model ($+0.4\%$ vs.\ $+0.9\%$ average accuracy) where the unique role of the softmax temperature makes tuning this hyperparameter more important.
Fig.~\ref{fig:acc_vs_temp_imagenet21k} shows this effect.
%the effect of the temperature on average accuracy, we see that the heteroscedastic model is more sensitive to the temperature hyperparameter than the homoscedastic model.

As described above (\S \ref{sec:proposed_model}), Fig.~\ref{fig:bias_variance_imagenet21k} shows the bias and variance throughout Imagenet-21k training, verifying our claim that the temperature plays the role of regulating a bias-variance trade-off. 

% \begin{figure*}[tbh]
%      \centering
%      \includegraphics[width=\textwidth]{dog_segs}
%      \caption{Image segmentation for an image from the internet (not in the PASCAL VOC dataset).
%      }
%      \label{fig:dog_segs}
% \end{figure*}

% %%%% Uncertainty
% \begin{figure*}[tbh]
%      \centering
%      \includegraphics[width=\textwidth]{dog_uncertainty.jpg}
%      \caption{Heatmaps of aleatoric uncertainty (per class variance) for two heteroscedastic models corresponding to $\tau=1.0$ (top) and the optimal temperature $\tau^*=0.05$ (bottom). $S=1000$.
%      }
%      \label{fig:uncertainty}
% \end{figure*}

\subsection{Image Segmentation}

% \begin{table*}[tbh]
% \centering
% \caption{Image segmentation results for heteroscedastic and homoscedastic models. $p$-values are from a paired sample two-tailed t-test where replicas are from corresponding random seeds. 25 replicates are used. T-tests are conducted w.r.t. the heteroscedastic mIoU at optimal $\tau^*$. The number of test server submissions is limited, so we only report p-values for the validation set. $^\ddagger$ p < 0.001.}
% \label{table:het_vs_hom_image_seg}
% \begin{tabular}{lc|ccc|ccc|}
% \toprule
%  &  & \multicolumn{3}{c|}{\textbf{Validation}} & \multicolumn{3}{c|}{\textbf{Test}} \\
% Dataset & $\tau^*$ & mIoU$_{\tau=\tau^*}$ & mIoU$_{\tau=1}$ & mIoU$_{\rm Hom}$ & mIoU$_{\tau=\tau^*}$ & mIoU$_{\tau=1}$ & mIoU$_{\rm Hom}$ \\
% \midrule
% Cityscapes & 0.05 & $\mathbf{75.32\%}^\ddagger$ & 74.22\% & 74.24\% & $\mathbf{77.35\%}$ & 76.36\% & 76.61\% \\ 
% PASCAL VOC & 0.05 & $\mathbf{85.89\%}^\ddagger$ & 84.55\% & 84.89\% & $\mathbf{84.65\%}$ & 83.93\% & 84.01\% \\ 
% \bottomrule
% \end{tabular}
% \end{table*}

\begin{table}[tbh]
\centering
\caption{Image segmentation results for heteroscedastic and homoscedastic models. Metrics are percentage mean intersection-over-union. $p$-values are from a paired sample two-tailed t-test, 25 replicates from corresponding random seeds are used. T-tests are conducted w.r.t. the heteroscedastic mIoU at optimal $\tau^*$. The number of test server submissions is limited, so we only report p-values for the validation set. $^\ddagger$ p < 0.001.}
\label{table:het_vs_hom_image_seg}
\begin{tabular}{lcccc}
\toprule
 Method & \multicolumn{2}{c}{\textbf{Cityscapes}} & \multicolumn{2}{c}{\textbf{PASCAL VOC}} \\
 & Valid & Test & Valid & Test \\
 \midrule
Homoscedastic & 74.24 & 76.61 & 84.89 & 84.01 \\
$\tau = 1.0$ \cite{kendall2017uncertainties} & 74.22 & 76.36 & 84.55 & 83.93 \\
Ours $\tau = \tau^*$ & $\mathbf{75.32}^\ddagger$ & $\mathbf{77.35}$ & $\mathbf{85.89}^\ddagger$ & $\mathbf{84.65}$ \\
\bottomrule
\end{tabular}
\end{table}

\begin{figure*}[tbh]
\centering
\includegraphics[width=0.95\textwidth]{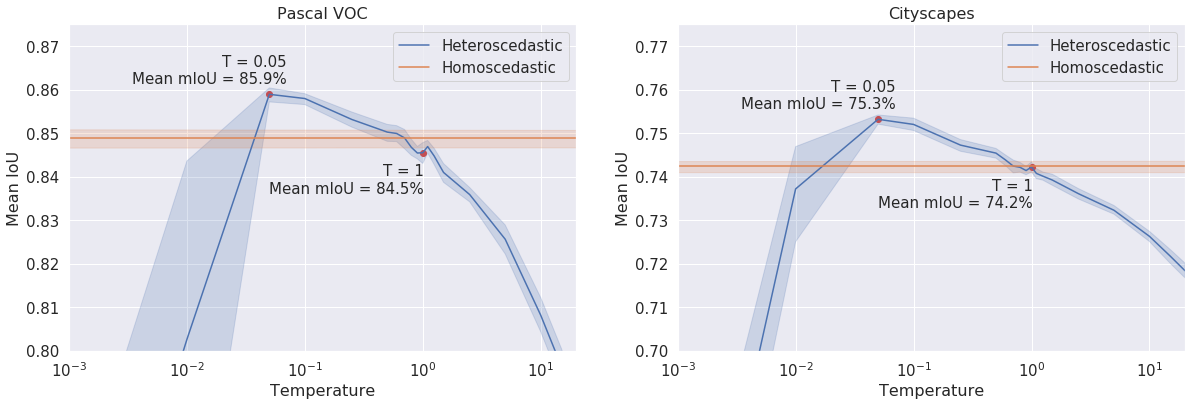}
\caption{Validation set image segmentation mIoU on PASCAL VOC and Cityscapes datasets. Results are averaged over 25 random seeds. The shaded area shows bootstrapped 95\% confidence intervals.}
\label{fig:image_seg_T_sweep}
\end{figure*}

Image segmentation datasets have naturally occurring heteroscedastic uncertainty. A single $512 \times 512$ image has $262,144$ pixels, so in practice human annotators cannot label pixels individually but label collections of pixels at a time. As a result annotations tend to be noisy at the boundaries of objects. We apply our heteroscedastic model to PASCAL VOC 2012 \cite{everingham2014pascal} and Cityscapes \cite{cordts2016cityscapes}, two popular image segmentation benchmarks. We follow the same end-to-end architecture and experimental setup as \citet{chen2018deeplabv3+}, see Appendix \ref{sec:imseg_arch_appendix} for details. Performance is measured by mean Intersection over Union (mIoU).

Fig.\ \ref{fig:image_seg_T_sweep} shows the effect of the softmax temperature on segmentation quality, using $S = 1000$ MC samples for the heteroscedastic method. Again we observe a classic trade-off curve with an optimal temperature in-between two extremes of bias and variance. Heteroscedastic models outperform the homoscedastic model for a range of temperatures. Similar to the controlled label noise experiments, the optimal temperature $\tau^\ast$ is not $1.0$. Furthermore, for both datasets, $\tau=1.0$ is outperformed by the \textit{homoscedastic} model on average. Table \ref{table:het_vs_hom_image_seg} shows that the differences in performance between the heteroscedastic model at the optimal temperature, the heteroscedastic model at $\tau = 1.0$ and the homoscedastic model are statistically significant. We report both validation set and test set results as the number of submissions to the test server is limited, which does not enable us to test the importance of the temperature parameter or compute p-values.

The difference in the models also leads to qualitatively different segmentations and uncertainty heatmaps. Fig. \ref{fig:dog_segs} shows an example segmentation, using the best homoscedastic, heteroscedastic at $\tau=1.0$ and at $\tau^\ast = 0.05$ models trained on PASCAL VOC. Reflecting the improvement in mean IoU the heteroscedastic segmentation at optimal temperature is qualitatively superior. Further examples are shown in Appendix \ref{sec:imseg_examples_appendix} where we have selected both success and failure cases.

Image segmentation provides a natural example of the additional advantages (other than improved predictive performance) of our method vs.\ other methods in the noisy labels literature. Our method also provides an estimate of aleatoric uncertainty. Fig.\ \ref{fig:uncertainty} demonstrates this, showing heat maps of per-class variance of the predictive distribution. As expected, the regions of highest aleatoric uncertainty are at object boundaries. Interestingly, the heteroscedastic uncertainty heatmaps at optimal temperature are more fine grained and precise than the $\tau = 1.0$ heatmaps.

\section{Conclusion}

% \RJ{revise narrative to remove ``true'', etc\dots}

Inspired by the econometrics literature, we have assumed a heteroscedastic latent variable generative process for noisy labels. We make a smooth approximation to the log-likelihood of this process with a temperature parameterized softmax. This approximation is equivalent in the zero temperature limit but in practice the temperature must be tuned to balance bias in the approximation and the variance of the gradients during training. The model of \citet{kendall2017uncertainties} can be viewed as a special case of our method with $\tau = 1.0$ and the choice of a Gaussian noise distribution.

We have shown improved log-likelihood, accuracy and calibration on image classification tasks with controlled label noise. Our method also provides significantly improved accuracy on the large-scale image classification benchmark Imagenet-21k which has naturally occurring label noise. On two image segmentation datasets, our method gives qualitatively and quantitatively improved segmentations over the state-of-the-art method.

\bibliography{references}

\clearpage

\onecolumn
\aistatstitle{A Simple Probabilistic Method for Deep Classification under Input-Dependent Label Noise: \\ 
Supplementary Materials}

\appendix

\section{Heteroscedastic Binary Classification}
\label{sec:binary_classification}

For multi-class classification we use the softmax as a smoothing function for the argmax with the guarantee of equivalence in a zero temperature limit. For binary classification it is more convenient to avoid the use of the vector valued argmax and softmax functions and simply have the model output the probability of one class being chosen, $p_1$, in which case the probability of the other class is simply $p_0 = 1 - p_1$:

\begin{equation}
\begin{split}
    p_1 &= P(u_{1}^{*} > u_{0}^{*}) \\
    &= P(u_{1}^{*} - u_{0}^{*} > 0) \\
    &= P(u^{*} > 0) \\
    &= \mathbb{E}_{\epsilon \sim \mathcal{N}(0, \sigma_{\mathbf{w}}(\mathbf{x}))} \brac{\mathds{1} \parencurly{u^{*} > 0}}\\
    &= \mathbb{E}_{\epsilon \sim \mathcal{N}(0, \sigma^{\mathbf{w}}(\mathbf{x}))} \brac{\lim_{\tau \to 0} \frac{1}{1 + \exp{(-u^{*}/\tau})}}\\
    &\approx \mathbb{E}_{\epsilon \sim \mathcal{N}(0, \sigma^{\mathbf{w}}(\mathbf{x}))} \brac{\frac{1}{1 + \exp{(-u^{*}/\tau})}}, \ \tau > 0
\end{split}
\label{eq:p_sigmoid}
\end{equation}

The key step is to replace the difference of the two latent variables $u_{1}^{*} - u_{0}^{*}$ with a single latent variable $u^{*}$ which is valid as all latent variables are members of the location-scale family $\mathcal{N}$. This sigmoid smoothing function has also been used in the econometrics literature \cite{train2009discrete}.

\newpage

\section{On the Connection Between MentorNet and Heteroscedastic Regression}
\label{sec:mentornet_connection}

The MentorNet \cite{MentorNet.2018} loss function takes the form:

\begin{equation}
    \frac{1}{n} \sum_{i=1}^{n} v_i \mathcal{L}(\hat{y}_i(\mathbf{x}_i; \mathbf{w}), y_i) + G(v_i)
\end{equation}

where  $v_i = g_m(\mathbf{x}_i ; \theta) \in [0, 1]$, is a per example weighting outputted by the neural network $g_m$. $G(v_i)$ is a regularizer to ensure that a weight of 0 is not used for all examples.

Note that for some choices of $G$ the optimal $g_m$ can be derived analytically when $\mathbf{w}$ is fixed. We have excluded the L2 penalty on the weights which can be added to the loss functions below for equivalence.

Supposing we make the standard choice of squared error loss function $\mathcal{L}(a, b) = (a - b)^2$ for regression, the MentorNet objective is:

\begin{equation}
    \frac{1}{n} \sum_{i=1}^{n} v_i (\hat{y}_i(\mathbf{x}_i; \mathbf{w}) - y_i)^2 + G(v_i)
\end{equation}

Taking the probabiltic approach, assuming a heteroscedastic Gaussian likelihood s.t.\ $y_i \sim \mathcal{N}(\hat{y}_i(\mathbf{x}_i; \mathbf{w}), \sigma^2(\mathbf{x}_i; \mathbf{w}'))$ the heteroscedastic objective is:

\begin{equation}
    \frac{1}{n} \sum_{i=1}^{n} \frac{1}{2} \exp(-m_i) (\hat{y}_i(\mathbf{x}_i; \mathbf{w}) - y_i)^2 + \frac{1}{2} m_i
\end{equation}

where $m_i = \log \sigma^2(\mathbf{x}_i; \mathbf{w}')$. Letting $v_i = \frac{1}{2} \exp(-m_i) \Rightarrow \frac{1}{2} m_i = -\log v_i - \log 2$ hence the MentorNet objective is recovered with the choice of $G(v_i) = -\log v_i - \log 2$. So for a particular choice of $G$ the MentorNet and heteroscedastic approaches are equivalent and in this case MentorNet has a satisfying probabilistic interpretation.

\section{Controlled Label Noise: Further Results}
\label{sec:more_synthetic_results}

\begin{figure}[tbh]
    \centering
    \includegraphics[width=\textwidth]{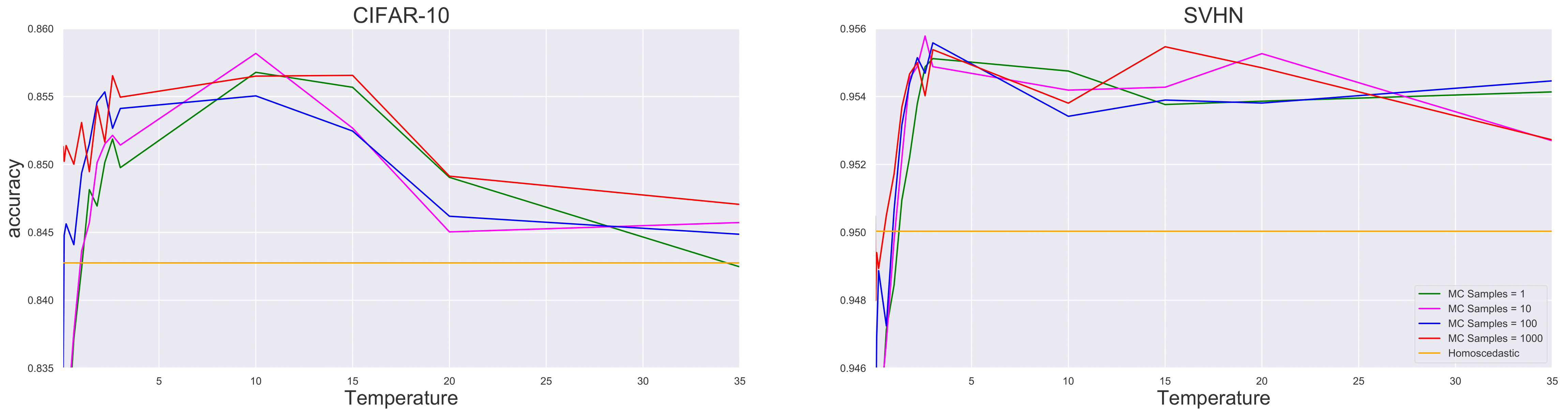}
    \caption{Effect of temperature and number of MC samples during training on \textit{noisy} test set accuracy.}
    \label{fig:synthetic_results_acc}
\end{figure}

\subsection{No Noise}
\label{sec:no_noise}

To test the generality of our method we evaluate it when trained on the original CIFAR-10 and SVHN datasets \textit{without} corrupted labels, see Table \ref{table:no_noise} for results. Interestingly, we observe that our method leads to higher test set log-likelihood and accuracy compared to the homoscedastic and \citet{kendall2017uncertainties} baselines. This demonstrates that 1) our method can be applied to datasets with labels that are considered to be clean and still lead to performance improvements and 2) these datasets may have a source heteroscedasticity e.g.\ from ambiguous or hard to label examples.

\npdecimalsign{.}
\npfourdigitnosep
\nprounddigits{3}
\begin{table}[tbh]
\setlength{\tabcolsep}{5pt}
\caption{No noise. CIFAR-10 and SVHN test dataset performance. Negative Log-likelihood (NLL) and expected calibration error (ECE) \cite{guo2017calibration} are computed on the noisy test set (with the same label corruption process applied to the training set). Clean accuracy (Acc) is computed on the test set with uncorrupted labels. For our method, optimal $\tau^\ast$ is determined by validation set log-likelihood. Number of samples $S = 1000$. $p$-values are from a paired sample two-tailed t-test with 25 replicates from corresponding random seeds. T-tests are conducted in reference to our method.}
\label{table:no_noise}
\centering
\begin{tabular}{lcccccc}
\toprule
Method &
  \multicolumn{3}{c}{CIFAR-10 ($\tau^\ast = 10$)} &
  \multicolumn{3}{c}{SVHN ($\tau^\ast = 10$)} \\
 & NLL & Acc & ECE & NLL & Acc & ECE \\
  \midrule
Homoscedastic &
  $\numprint{0.39351628399999994}^{\dagger}$ &
  $\numprint{0.8873200000000001}$ &
  $\numprint{0.05028130228}^{\dagger}$ &
  $\numprint{0.156381622}^{\ddagger}$ &
  $\numprint{0.9610663696}^{\ddagger}$ &
  $\numprint{0.013349838328000001}$ \\
$\tau = 1.0$ \cite{kendall2017uncertainties} &
  $\numprint{0.397947596}^{\ddagger}$ &
  $\mathbf{\numprint{0.8922079999999998}}$ &
  $\numprint{0.05389260844000001}^{\ddagger}$ &
  $\numprint{0.15879492320000002}^{\ddagger}$ &
  $\numprint{0.9605224388000001}^{\ddagger}$ &
  $\numprint{0.013586439427999999}$ \\
Ours $\tau = \tau^\ast$ &
  $\mathbf{\numprint{0.37456327480000007}}$ &
  $\mathbf{\numprint{0.8918640000000001}}$ &
  $\mathbf{\numprint{0.04332951231999999}}$ &
  $\mathbf{\numprint{0.1453456956}}$ &
  $\mathbf{\numprint{0.9632267996000001}}$ &
  $\mathbf{\numprint{0.012297000500000002}}$ \\
\bottomrule
\end{tabular}
\begin{flushleft}
\small
 $^*$ p < 0.05 \hspace{5mm}
 $^\dagger$ p < 0.01  \hspace{5mm}
 $^\ddagger$ p < 0.001
\end{flushleft}
\end{table}

\subsection{Uniform Noise}
\label{sec:uniform_noise}

We evaluate our method and all baselines on the image classification datasets under uniform/homoscedastic noise, see Table \ref{table:uniform_noise}. We randomly reassign 20\% of labels to a label in 1-10 (with equal probability for each label). Our method performs competitively on this benchmark and outperforms the homoscedastic and $\tau = 1.0$ baselines. However the MentorNet and Co-teaching benchmarks have higher accuracy under uniform noise. We note that in many applications the source of label noise is complex and likely input dependent and that simple scenarios such as uniform label noise may be unrealistic in practice.

\npdecimalsign{.}
\npfourdigitnosep
\nprounddigits{3}
\begin{table}[tbh]
\setlength{\tabcolsep}{5pt}
\caption{Uniform/homoscedastic noise. CIFAR-10 and SVHN test dataset performance. Negative Log-likelihood (NLL) and expected calibration error (ECE) \cite{guo2017calibration} are computed on the noisy test set (with the same label corruption process applied to the training set). Clean accuracy (Acc) is computed on the test set with uncorrupted labels. For our method, optimal $\tau^\ast$ is determined by validation set log-likelihood. Number of samples $S = 1000$. $p$-values are from a paired sample two-tailed t-test with 25 replicates from corresponding random seeds. T-tests are conducted in reference to our method.}
\label{table:uniform_noise}
\centering
\begin{tabular}{lcccccc}
\toprule
Method &
  \multicolumn{3}{c}{CIFAR-10 ($\tau^\ast = 3$)} &
  \multicolumn{3}{c}{SVHN ($\tau^\ast = 10$)} \\
 & NLL & Acc & ECE & NLL & Acc & ECE \\
  \midrule
Homoscedastic &
  $\numprint{1.215470044}^{\ddagger}$ &
  $\numprint{0.8384160000000002}^{\ddagger}$ &
  $\numprint{0.05334821323999999}^{\dagger}$ &
  $\numprint{0.9852078688}^{\ddagger}$ &
  $\numprint{0.950030734}^{\ddagger}$ &
  $\numprint{0.042446814039999996}^{\ddagger}$ \\
$\tau = 1.0$ \cite{kendall2017uncertainties} &
  $\numprint{1.188885564}^{*}$ &
  $\numprint{0.845704}$ &
  $\numprint{0.045090009359999995}$ &
  $\numprint{0.9753131312000002}^{\ddagger}$ &
  $\numprint{0.9517240360000001}^{\ddagger}$ &
  $\numprint{0.027889788640000003}^{\ddagger}$ \\
Ours $\tau = \tau^\ast$ &
  $\mathbf{\numprint{1.170788444}}$ &
  $\numprint{0.849384}$ &
  $\mathbf{\numprint{0.04439290412}}$ &
  $\mathbf{\numprint{0.9626653507999999}}$ &
  $\numprint{0.9554671156000001}$ &
  $\mathbf{\numprint{0.023161711639999995}}$ \\
  \midrule
Bootstrapping &
  $\numprint{1.2289858919999999}^{\ddagger}$ &
  $\numprint{0.8456880000000001}$ &
  $\numprint{0.0765357148}^{\ddagger}$ &
  $\numprint{0.9822940132000001}^{\ddagger}$ &
  $\numprint{0.9532606016000001}^{*}$ &
  $\numprint{0.04977747279999999}^{\ddagger}$\\    
MentorNet &
  $\numprint{2.4250645480000004}^{\ddagger}$ &
  $\numprint{0.8571040000000001}^{\dagger}$ &
  $\numprint{0.22579973239999998}^{\ddagger}$ &
  $\numprint{1.639157924}^{\ddagger}$ &
  $\numprint{0.9561339900000001}^{*}$ &
  $\numprint{0.15094565606000001}^{\ddagger}$ \\
Co-teaching &
  $\numprint{2.6268767279999996}^{\ddagger}$ &
  $\mathbf{\numprint{0.864848}}^{\ddagger}$ &
  $\numprint{0.21544865}^{\ddagger}$ &
  $\numprint{2.239157924}^{\ddagger}$ &
  $\mathbf{\numprint{0.9580086040000001}}^{\ddagger}$ &
  $\numprint{0.19070374}^{\ddagger}$ \\
\bottomrule
\end{tabular}
\begin{flushleft}
\small
 $^*$ p < 0.05 \hspace{5mm}
 $^\dagger$ p < 0.01  \hspace{5mm}
 $^\ddagger$ p < 0.001
\end{flushleft}
\end{table}

\subsection{Varying Noise Level}
\label{sec:varying_noise}

We vary the level of heteroscedastic noise to see the effect of using our model when there is a small/large amount of heteroscedastic noise relative to the results in the paper. Table \ref{table:reduced_hetero} shows the results for a reduced level of heteroscedastic noise. In the main paper results are presented where labels 1-4 are left uncorrupted while labels 5-10 have 10\%, 20\%, 30\%, 40\%, 50\% and 60\% probability of being redrawn from a uniform distribution over all labels 1-10. For the results shown in Table \ref{table:reduced_hetero}, labels 1-4 are also left uncorrupted but the probability of corruption for the other labels is halved, so labels 5-10 have 5\%, 10\%, 15\%, 20\%, 25\% and 30\% probability of being redrawn from a uniform distribution over all labels 1-10.

\npdecimalsign{.}
\npfourdigitnosep
\nprounddigits{3}
\begin{table}[tbh]
\setlength{\tabcolsep}{5pt}
\caption{\textbf{Reduced} heteroscedastic noise. CIFAR-10 and SVHN test dataset performance. Negative Log-likelihood (NLL) and expected calibration error (ECE) \cite{guo2017calibration} are computed on the noisy test set (with the same label corruption process applied to the training set). Clean accuracy (Acc) is computed on the test set with uncorrupted labels. For our method, optimal $\tau^\ast$ is determined by validation set log-likelihood. Number of samples $S = 1000$. $p$-values are from a paired sample two-tailed t-test with 25 replicates from corresponding random seeds. T-tests are conducted in reference to our method.}
\label{table:reduced_hetero}
\centering
\begin{tabular}{lcccccc}
\toprule
Method &
  \multicolumn{3}{c}{CIFAR-10 ($\tau^\ast = 3$)} &
  \multicolumn{3}{c}{SVHN ($\tau^\ast = 10$)} \\
 & NLL & Acc & ECE & NLL & Acc & ECE \\
  \midrule
Homoscedastic &
  $\numprint{0.848376108}^{\ddagger}$ &
  $\numprint{0.8640320000000001}$ &
  $\numprint{0.0456583554}$ &
  $\numprint{0.4684606936000001}^{\ddagger}$ &
  $\numprint{0.9557498527999999}^{\ddagger}$ &
  $\numprint{0.016115140439999998}^{\ddagger}$ \\
$\tau = 1.0$ \cite{kendall2017uncertainties} &
  $\numprint{0.8580820912}^{\ddagger}$ &
  $\numprint{0.8700239999999999}^{\ddagger}$ &
  $\numprint{0.05237877084}^{*}$ &
  $\numprint{0.4628090731999999}^{\ddagger}$ &
  $\numprint{0.9572526136}^{\ddagger}$ &
  $\numprint{0.013051493192000002}$ \\
Ours $\tau = \tau^\ast$ &
  $\mathbf{\numprint{0.8126940216000003}}$ &
  $\numprint{0.8713199999999999}$ &
  $\mathbf{\numprint{0.043106302679999986}}$ &
  $\mathbf{\numprint{0.4516730956}}$ &
  $\numprint{0.9599538991999998}$ &
  $\mathbf{\numprint{0.012100682799999998}}$ \\
  \midrule
Bootstrapping &
  $\numprint{0.8537609723999999}^{\ddagger}$ &
  $\numprint{0.869608}$ &
  $\numprint{0.06504927288}^{\ddagger}$ &
  $\numprint{0.46748500480000005}^{\ddagger}$ &
  $\numprint{0.9578457292}^{\ddagger}$ &
  $\numprint{0.02325561972}^{\ddagger}$\\    
MentorNet &
  $\numprint{1.528587116}^{\ddagger}$ &
  $\numprint{0.875712}^{*}$ &
  $\numprint{0.134715429}^{\ddagger}$ &
  $\numprint{1.1036018048}^{\ddagger}$ &
  $\numprint{0.9607037455999999}$ &
  $\numprint{0.08870061552000001}^{\ddagger}$ \\
Co-teaching &
  $\numprint{1.5603855400000004}^{\ddagger}$ &
  $\mathbf{\numprint{0.877448}}^{\ddagger}$ &
  $\numprint{0.13830194815999997}^{\ddagger}$ &
  $\numprint{1.085654436}^{\ddagger}$ &
  $\mathbf{\numprint{0.961874616}}^{\ddagger}$ &
  $\numprint{0.08545438868000002}^{\ddagger}$ \\
\bottomrule
\end{tabular}
\begin{flushleft}
\small
 $^*$ p < 0.05 \hspace{5mm}
 $^\dagger$ p < 0.01  \hspace{5mm}
 $^\ddagger$ p < 0.001
\end{flushleft}
\end{table}

Table \ref{table:increased_hetero} shows the results for an increased level of heteroscedastic noise. For the results shown in Table \ref{table:increased_hetero}, labels 1-10 have 20\%, 25\%, ..., 65\% probability of being redrawn from a uniform distribution over all labels 1-10. With this increased level of heteroscedastic noise, the magnitude of the performance improvement from our method is increased relative to the reduced levels of heteroscedastic noise in the results in Table \ref{table:reduced_hetero} and Table 1 in the main paper.

\npdecimalsign{.}
\npfourdigitnosep
\nprounddigits{3}
\begin{table}[tbh]
\setlength{\tabcolsep}{5pt}
\caption{\textbf{Increased} heteroscedastic noise. CIFAR-10 and SVHN test dataset performance. Negative Log-likelihood (NLL) and expected calibration error (ECE) \cite{guo2017calibration} are computed on the noisy test set (with the same label corruption process applied to the training set). Clean accuracy (Acc) is computed on the test set with uncorrupted labels. For our method, optimal $\tau^\ast$ is determined by validation set log-likelihood. Number of samples $S = 1000$. $p$-values are from a paired sample two-tailed t-test with 25 replicates from corresponding random seeds. T-tests are conducted in reference to our method.}
\label{table:increased_hetero}
\centering
\begin{tabular}{lcccccc}
\toprule
Method &
  \multicolumn{3}{c}{CIFAR-10 ($\tau^\ast = 3$)} &
  \multicolumn{3}{c}{SVHN ($\tau^\ast = 10$)} \\
 & NLL & Acc & ECE & NLL & Acc & ECE \\
  \midrule
Homoscedastic &
  $\numprint{1.7769560080000002}^{\ddagger}$ &
  $\numprint{0.7658400000000001}^{\ddagger}$ &
  $\numprint{0.04723909632000001}^{*}$ &
  $\numprint{1.479871444}^{\ddagger}$ &
  $\numprint{0.9298832188}^{\ddagger}$ &
  $\numprint{0.03470021544000001}^{\ddagger}$ \\
$\tau = 1.0$ \cite{kendall2017uncertainties} &
  $\numprint{1.7520816879999999}^{\ddagger}$ &
  $\numprint{0.7832799999999999}^{\ddagger}$ &
  $\mathbf{\numprint{0.04023203628}}$ &
  $\numprint{1.467552608}^{\ddagger}$ &
  $\numprint{0.9337461572000001}^{\ddagger}$ &
  $\numprint{0.021226610400000005}^{\ddagger}$ \\
Ours $\tau = \tau^\ast$ &
  $\mathbf{\numprint{1.723935012}}$ &
  $\mathbf{\numprint{0.806512}}$ &
  $\numprint{0.04125051807999999}$ &
  $\mathbf{\numprint{1.4523487160000002}}$ &
  $\mathbf{\numprint{0.9419483679999999}}$ &
  $\mathbf{\numprint{0.017340620539999997}}$ \\
  \midrule
Bootstrapping &
  $\numprint{1.794873632}^{\ddagger}$ &
  $\numprint{0.7807520000000001}^{\dagger}$ &
  $\numprint{0.09988820423999999}^{\ddagger}$ &
  $\numprint{1.4839819280000004}^{\ddagger}$ &
  $\numprint{0.9389797275999999}^{*}$ &
  $\numprint{0.0764061}^{\ddagger}$\\
MentorNet &
  $\numprint{4.478599004}^{\ddagger}$ &
  $\numprint{0.8037920000000001}^{*}$ &
  $\numprint{0.3963202964}^{\ddagger}$ &
  $\numprint{2.0614444520000004}^{\ddagger}$ &
  $\numprint{0.936272276}^{\ddagger}$ &
  $\numprint{0.11167248588000003}^{\ddagger}$ \\
Co-teaching &
  $\numprint{5.270634835999999}^{\ddagger}$ &
  $\mathbf{\numprint{0.806512}}$ &
  $\numprint{0.4056581223999999}^{\ddagger}$ &
  $\numprint{4.610668548000001}^{\ddagger}$ &
  $\numprint{0.9178641672000002}^{\ddagger}$ &
  $\numprint{0.36004710880000007}^{\ddagger}$ \\
\bottomrule
\end{tabular}
\begin{flushleft}
\small
 $^*$ p < 0.05 \hspace{5mm}
 $^\dagger$ p < 0.01  \hspace{5mm}
 $^\ddagger$ p < 0.001
\end{flushleft}
\end{table}

\subsection{Platt-scaling Ablation}
\label{sec:platt_scaling}

Table \ref{table:hetero_vs_homo_full} shows the full Platt-scaling ablation results compared to Table \ref{table:hetero_vs_homo} in the main paper. All methods benefit from Platt-scaling. However, our method remains best or joint best across all metrics when Platt-scaling is applied to all methods. 

\npdecimalsign{.}
\npfourdigitnosep
\nprounddigits{3}
\begin{table}[tbh]
\setlength{\tabcolsep}{5pt}
\caption{CIFAR-10 and SVHN test dataset performance. Negative Log-likelihood (NLL) and expected calibration error (ECE) \cite{guo2017calibration} are computed on the noisy test set (with the same label corruption process applied to the training set). Clean accuracy (Acc) is computed on the test set with uncorrupted labels. For our method, optimal $\tau^\ast$ is determined by validation set log-likelihood. Number of samples $S = 1000$. $p$-values are from a paired sample two-tailed t-test with 25 replicates from corresponding random seeds. T-tests are conducted in reference to our method.}
\label{table:hetero_vs_homo_full}
\centering
\begin{tabular}{lcccccc}
\toprule
Method &
  \multicolumn{3}{c}{CIFAR-10 ($\tau^\ast = 3$)} &
  \multicolumn{3}{c}{SVHN ($\tau^\ast = 10$)} \\
 & NLL & Acc & ECE & NLL & Acc & ECE \\
  \midrule
Homoscedastic &
  $\numprint{1.1378800120000003}^{\ddagger}$ &
  $\numprint{0.8427520000000003}^{\ddagger}$ &
  $\numprint{0.04342196612}$ &
  $\numprint{0.6886919395999999}^{\ddagger}$ &
  $\numprint{0.950030734}^{\ddagger}$ &
  $\numprint{0.01811325112}^{\dagger}$ \\
\ \ + Platt-scaling &
  $\numprint{1.1242398320000002}^{\ddagger}$ &
  $\numprint{0.8427520000000003}^{\ddagger}$ &
  $\numprint{0.030100399799999997}^{\ddagger}$ &
  $\numprint{0.6856170852}^{\ddagger}$ &
  $\numprint{0.950030734}^{\ddagger}$ &
  $\numprint{0.01523184818}^{\ddagger}$ \\
$\tau = 1.0$ \cite{kendall2017uncertainties} &
  $\numprint{1.1290426839999999}^{\ddagger}$ &
  $\numprint{0.853088}^{\dagger}$ &
  $\numprint{0.049608294080000004}^{*}$ &
  $\numprint{0.6800222575999999}^{\ddagger}$ &
  $\numprint{0.9517240360000001}^{\ddagger}$ &
  $\numprint{0.014971369119999998}$ \\
\ + Platt.\ &
  $\numprint{1.111916096}^{\ddagger}$ &
  $\numprint{0.853088}^{\dagger}$ &
  $\numprint{0.02576470008}^{*}$ &
  $\numprint{0.6780429372000001}^{\ddagger}$ &
  $\numprint{0.9517240360000001}^{\ddagger}$ &
  $\numprint{0.013134200620000001}^{*}$ \\
Ours $\tau = \tau^\ast$ &
  $\mathbf{\numprint{1.098491052}}$ &
  $\mathbf{\numprint{0.858624}}$ &
  $\mathbf{\numprint{0.04191602044000001}}$ &
  $\mathbf{\numprint{0.6689119888}}$ &
  $\mathbf{\numprint{0.9554671156000001}}$ &
  $\mathbf{\numprint{0.013474776364000003}}$ \\
\ + Platt.\ &
  $\mathbf{\numprint{1.0862124160000004}}$ &
  $\mathbf{\numprint{0.858624}}$ &
  $\mathbf{\numprint{0.021780285459999996}}$ &
  $\mathbf{\numprint{0.6688605748}}$ &
  $\mathbf{\numprint{0.9554671156000001}}$ &
  $\mathbf{\numprint{0.0110610903}}$ \\
  \midrule
Bootstrapping &
  $\numprint{1.1569313920000002}^{\ddagger}$ &
  $\numprint{0.8509359999999999}^{\ddagger}$ &
  $\numprint{0.07696113196}^{\ddagger}$ &
  $\numprint{0.6879735243999999}^{\ddagger}$ &
  $\numprint{0.9527719783999999}^{\ddagger}$ &
  $\numprint{0.03523115396}^{\ddagger}$\\    
\ \ + Platt-scaling &
  $\numprint{1.1215284280000002}^{\ddagger}$ &
  $\numprint{0.8509359999999999}^{\ddagger}$ &
  $\numprint{0.034695889439999995}^{\ddagger}$ &
  $\numprint{0.6776840907999998}^{\ddagger}$ &
  $\numprint{0.9527719783999999}^{\ddagger}$ &
  $\mathbf{\numprint{0.011247024875999998}}$ \\
MentorNet &
  $\numprint{1.7775430679999997}^{\ddagger}$ &
  $\numprint{0.8576960000000001}$ &
  $\numprint{0.14402864735999998}^{\ddagger}$ &
  $\numprint{1.3733280599999997}^{\ddagger}$ &
  $\mathbf{\numprint{0.9554931212000001}}$ &
  $\numprint{0.10695505087999999}^{\ddagger}$ \\
\ \ + Platt-scaling &
  $\numprint{1.5958343599999998}^{\ddagger}$ &
  $\numprint{0.8576960000000001}$ &
  $\numprint{0.1326711428}^{\ddagger}$ &
  $\numprint{1.0503218928}^{\ddagger}$ &
  $\mathbf{\numprint{0.9554931212000001}}$ &
  $\numprint{0.09570506232000003}^{\ddagger}$ \\
Co-teaching &
  $\numprint{2.506220476}^{\ddagger}$ &
  $\numprint{0.8568159999999999}$ &
  $\numprint{0.21968032959999997}^{\ddagger}$ &
  $\numprint{1.9085762600000005}^{\ddagger}$ &
  $\numprint{0.9541425995999999}$ &
  $\numprint{0.151293794}^{\ddagger}$ \\
\ \ + Platt-scaling &
  $\numprint{2.157522484}^{\ddagger}$ &
  $\numprint{0.8568159999999999}$ &
  $\numprint{0.20121231439999998}^{\ddagger}$ &
  $\numprint{1.3556756719999996}^{\ddagger}$ &
  $\numprint{0.9541425995999999}$ &
  $\numprint{0.13891620064000001}^{\ddagger}$ \\
\bottomrule
\end{tabular}
\begin{flushleft}
\small
 $^*$ p < 0.05 \hspace{5mm}
 $^\dagger$ p < 0.01  \hspace{5mm}
 $^\ddagger$ p < 0.001
\end{flushleft}
\end{table}

\section{Controlled Label Noise Experiments: Architectural and Training Details}
\label{sec:synthetic_appendix}

For all experiments on CIFAR-10 and SVHN datasets we use a similar architecture to \citet{CoTeaching.2018}. See Table \ref{tab:controlled_noise_architecture} for details. We make one change, replacing the use of Batch Normalization \cite{ioffe2015batch} with Group Normalization \cite{wu2018group} with 2 groups. The slopes of Leaky ReLU activation functions are set to 0.01.

\begin{table}
	\centering
	\caption{Covolutional network architectures used in experiments on MNIST, CIFAR-10 and SVHN datasets \cite{CoTeaching.2018}.}
	\label{tab:controlled_noise_architecture}
	\begin{tabular}{c}
		3$\times$3 conv 128 filters, group norm, LReLU   \\
		3$\times$3 conv 128 filters, group norm, LReLU   \\
		3$\times$3 conv 128 filters, group norm, LReLU  \\ \hline
		2$\times$2 max-pool, stride 2 \\
		dropout, $p=0.25$ \\ \hline
		3$\times$3 conv 256 filters, group norm, LReLU  \\
		3$\times$3 conv 256 filters, group norm, LReLU   \\
		3$\times$3 conv 256 filters, group norm, LReLU  \\ \hline
		2$\times$2 max-pool, stride 2 \\
		dropout, $p=0.25$ \\ \hline
		3$\times$3 conv 512 filters, group norm, LReLU \\
		3$\times$3 conv 256 filters, group norm, LReLU \\
		3$\times$3 conv 128 filters, group norm, LReLU  \\ \hline
		average-pool \\ \hline
        $f^{\mathbf{w}}(\mathbf{x})$: dense 128$\rightarrow$10, $\sigma^{\mathbf{w}}(\mathbf{x})$: dense 128$\rightarrow$10, softplus \\ \hline
	\end{tabular}
\end{table}

We train with Adam with default parameters; learning rate $ = 0.001$, $\beta_1 = 0.9$, $\beta_2 = 0.999$, $\epsilon=1e-07$. Networks are trained for a maximum of 1,000 epochs, being stopped early if validation set accuracy has not improved in 10 epochs. The best validation set checkpoint is used for test set evaluation.

CIFAR-10 images are 32x32 colour images. We use the standard 10,000 CIFAR-10 test images, we use 10,000 of the 50,000 CIFAR-10 training examples as a validation set and the remaining 40,000 as the training set. SVHN images are also 32x32 colour images. We use the standard 26,032 test images and 10,000 of the images in the standard training set split as a validation set with the remaining 63,257 used as a training set. All images are scaled to [0, 1] by dividing elementwise by 255.

For heteroscedastic models we search over the following temperatures: 0.025, 0.05, 0.1, 0.2, 0.6, 1.0, 1.4, 1.8, 2.2, 2.6, 3.0, 5.0, 10.0, 15.0, 20.0, 35.0, 50.0, 100.0, picking the optimal temperature on the validation set.

For the Co-teaching baseline \cite{CoTeaching.2018} we assume knowledge of the percentage of mis-labelled examples in the dataset and use this as the noise rate in the Co-teaching method. Similarly to the Co-teaching paper we allow $T_k = 10$, the number of epochs to linearly increase cutoff $R(T)$.

For the Bootstrap baseline \cite{reed6596training}, as per the original paper we set $\beta = 0.8$.

For the MentorNet baseline \cite{MentorNet.2018} we search over the following values for $\lambda_2$: 0.0, 0.5, 1.0, 2.0 and over 0, 5 and 10 epochs for the number of burn-in epochs. We set the decay factor to update the loss moving average to 0.9. We again assume knowledge of the percentage of mis-labelled examples in the dataset and use 1.0 - the noise rate as the quantile to determine the loss value to update the exponential moving average estimate of $\lambda_1$.

\section{Image Segmentation}
\label{sec:imseg_appendix}
\subsection{Architecture and Training Details}
\label{sec:imseg_arch_appendix}

We replicate the \texttt{DeepLabv3+} \cite{chen2018deeplabv3+} architecture and training setup which achieves state-of-the-art image segmentation results. \texttt{DeepLabv3+} uses an Xception \cite{chollet2017xception} based architecture with an added decoder module introduced. In particular we use the Xception65 architecture with an output stride of $16$ \cite{chen2018deeplabv3+}. The \texttt{DeepLabv3+} method has three training stages; pretraining on JFT and MSCoco, followed by a training phase with output stride 16 (on augmented and/or coarsely labelled data) during which batch norm parameters are fine tuned. Finally, they train for 30K steps using the SGD optimizer with learning rate of $0.001$ and otherwise default parameters. For experimentation, this final stage is trained at output stride 16, but models evaluated on the test server are trained with output stride 8 on the training and validation data.
We warm start all our models from the end of the second training phase, and attempt to replicate the training set up discussed. We initially train on the train set and the fine-tune temperatures on the validation set; we then repeat this stage using the training \& validation data using the best parameter settings from the validation set, and evaluate that on the test server.

In the homoscedastic model, a single convolution is applied to the output of a decoder, followed by bilinear upsampling to the size of the image, in order to compute logits for each pixel. For the heteroscedastic model, strictly speaking, either the final features should be upsampled to the original image size and used to compute correctly sized scale and location parameters, or the scale and location parameters should be computed at a lower dimension and upsampled to full size. However, this increases the number of MC samples required (by a factor of output stride), which makes it difficult to fit in the memory on a single device. We therefore sample the ``logits'' at a lower dimension and upsample to full image dimensions via bilinear interpolation.

\label{sec:imseg_test_appendix}

% \onecolumn
\newpage
\subsection{Image Segmentation Examples}
\label{sec:imseg_examples_appendix}

% \vspace{-7cm}

\begin{flushleft}
\seg{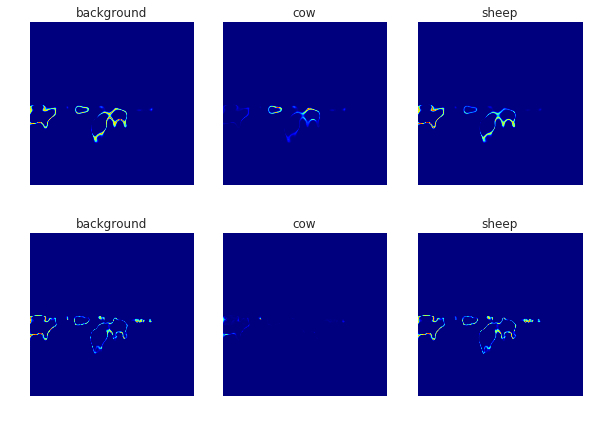}
\seg{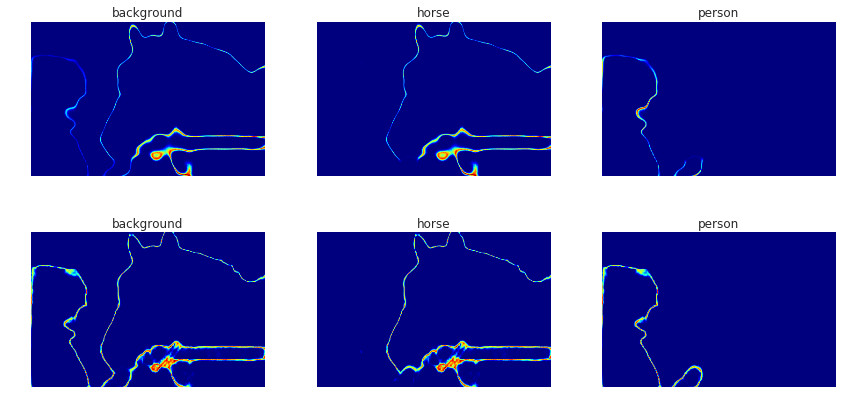}
\seg{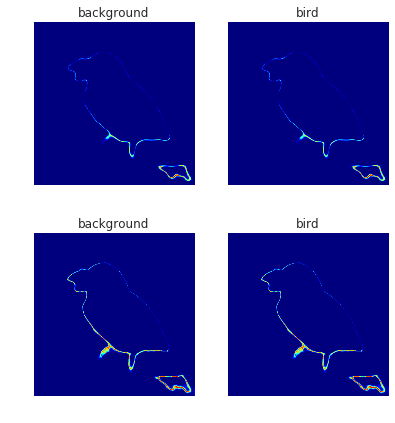}
\newpage
\seg{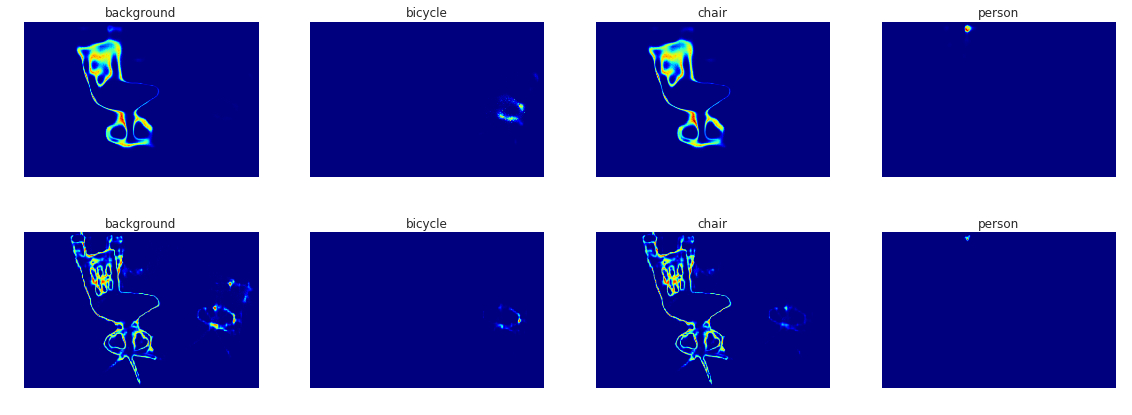}
\seg{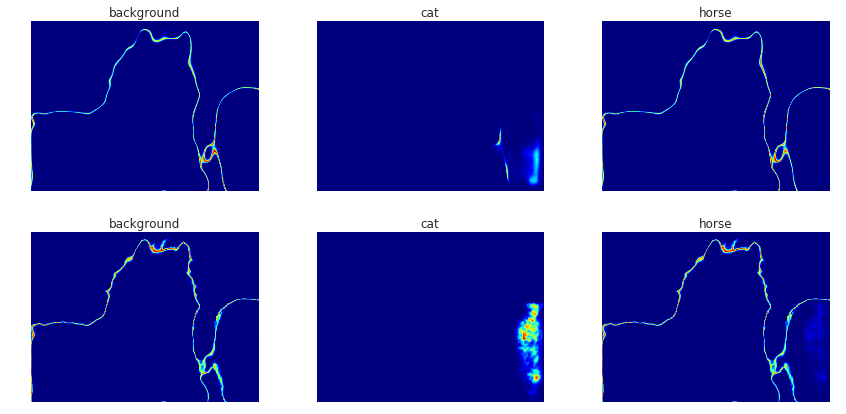}
\seg{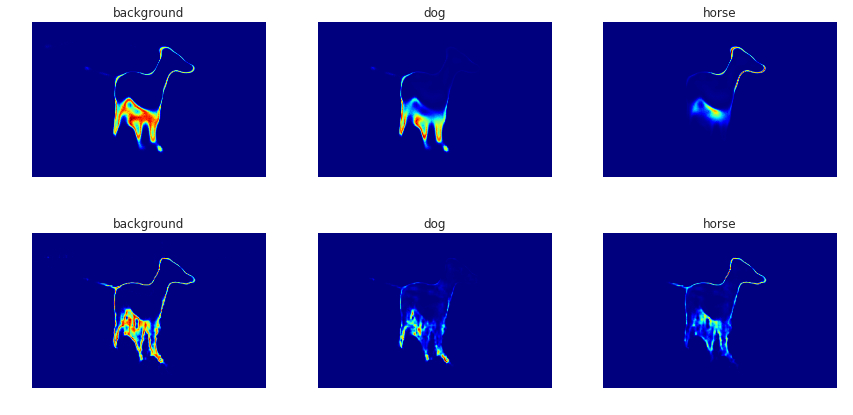}
\newpage
\seg{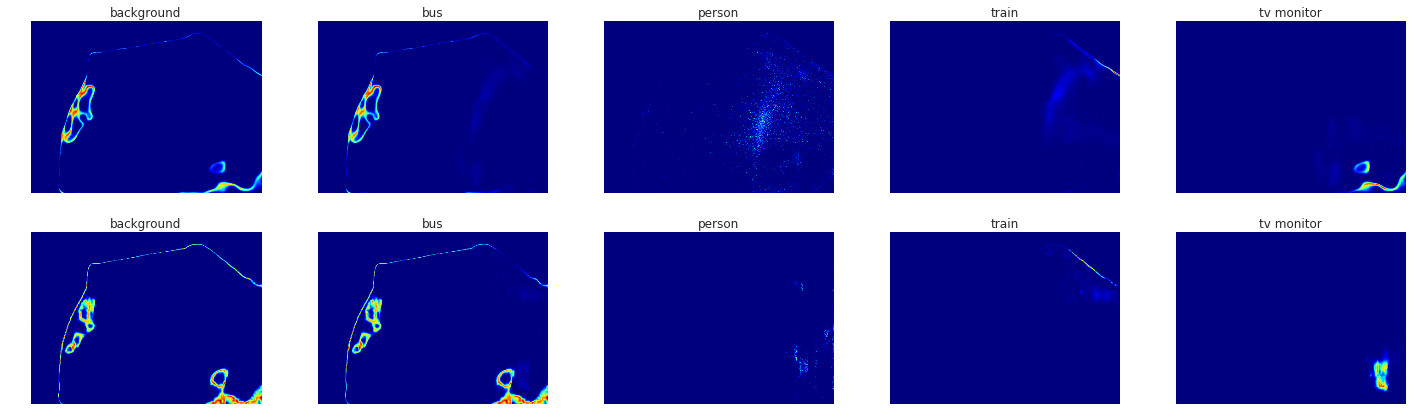}
\seg{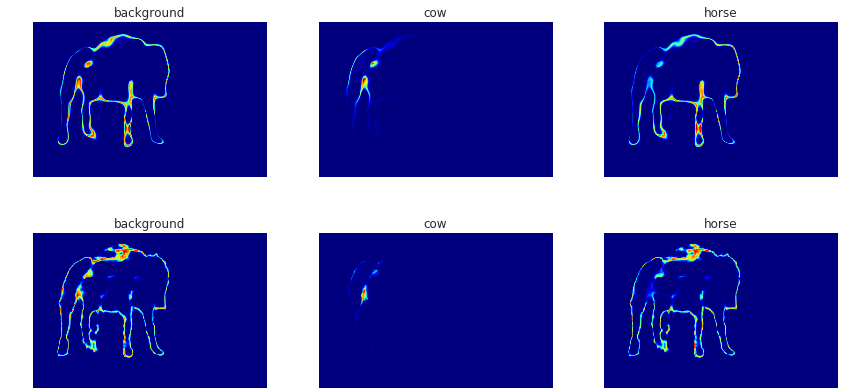}
\seg{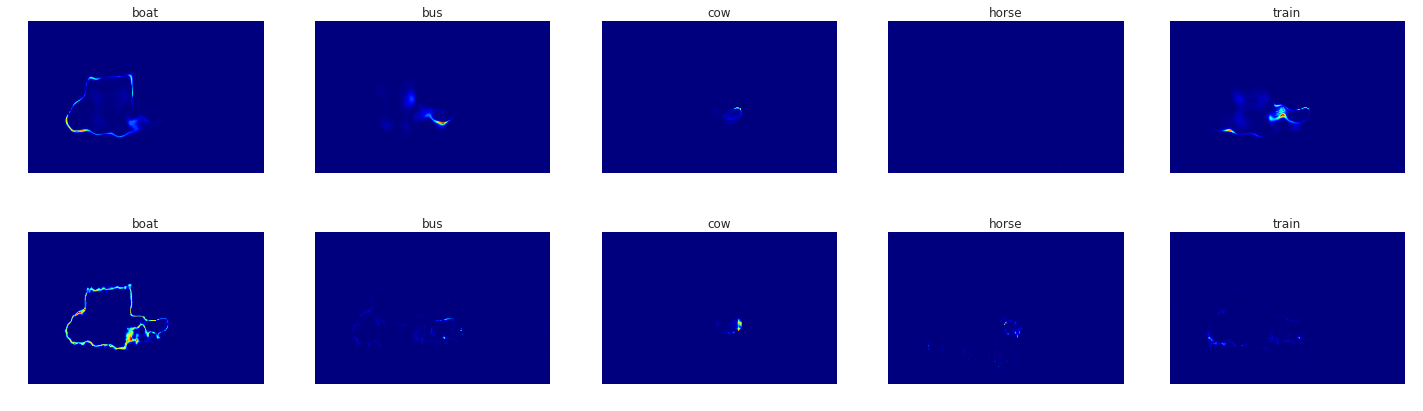}
\end{flushleft}
% \twocolumn

\end{document}